\documentclass[letterpaper, 10 pt, conference]{ieeeconf}  

\usepackage{xcolor}
\usepackage{graphicx}

\IEEEoverridecommandlockouts                              

\title{\LARGE \bf
Toward Humanoid Robots in Construction: A Teleoperation Feasibility Study
}

\author{Parastoo Ali Pour$^{1}$, David R. Martin$^{2}$, Chang Min Hur$^{3}$, Bo Zhang$^{3}$, Tommy Zhou$^{2}$, Brandon Thomas Lichter$^{2}$,  \\ Shane Stanfield$^{4}$, Pramod Khargonekar$^{2}$, Mohammad Abdullah Al Faruque$^{2}$
\thanks{This work was supported in part by Genova Construction.}
\thanks{$^{1}$ The author is with the Department of Mechanical and Aerospace Engineering, University of California at Irvine, Irvine, CA 92697 USA
        {\tt\small alipourp@uci.edu}}%
\thanks{$^{2}$The authors are with the Department of Electrical Engineering and Computer Science, University of California at Irvine, Irvine, CA 92697 USA
        {\tt\small \{davidrm3, tommyz8, blichter, pramod.khargonekar, alfaruqu\}@uci.edu}}%
\thanks{$^{3}$The authors are with the Professional Master of Embedded and Cyber-physical Systems Program, University of California at Irvine, Irvine, CA 92697 USA
        {\tt\small \{hurcm, boz18\}@uci.edu}}%
\thanks{$^{4}$The author is with Genova Construction (https://genovacd.com/).
        {\tt\small shane@genovacd.com}}%
}

\begin{document}

\maketitle
\thispagestyle{empty}
\pagestyle{empty}

\begin{abstract}
We present a teleoperation system that enables a single operator to perform construction tasks on a Unitree G1 humanoid, combining extended reality (XR) based upper body control with pedal-based locomotion to enable simultaneous manipulation and locomotion. Motivated by persistent labor shortages, hazardous working conditions, and challenges in humanoid autonomy, we investigate teleoperation as a practical near-term approach for reducing physical strain on workers while generating high quality demonstration data. We evaluate the system on two representative construction tasks drawn from O*NET occupational database, and report task success and completion time relative to a manual baseline. The system achieved 100\% success on tool transport and 80\% success on surface painting, with teleoperation requiring substantially more time compared to manual execution. 

\end{abstract}

\section{INTRODUCTION}

Construction remains one of the least automated sectors despite persistent labor shortages, hazardous working conditions, and declining productivity \cite{c1}. This stands in sharp contrast to more structured environments like manufacturing \cite{c2} and warehousing \cite{azadeh2019robotized}, where advances in robotics and artificial intelligence have automated many core tasks. A key barrier to automating construction labor is the dynamic and unstructured nature of construction sites, requiring robots that navigate human-scale spaces and perform diverse tasks that combine mobility with dexterous manipulation \cite{c1}. Existing construction automation has largely focused on task-specific systems, either in controlled, off-site environments where robots perform repetitive tasks such as frame prefabrication \cite{c3} \cite{brosque}, or directly on construction sites for activities such as bricklaying, drywall installation, and robotic assembly \cite{brosque} \cite{onsite}. This motivates the development of robotic systems that can operate directly in construction sites while supporting a broader range of tasks.

\begin{figure}[t]
    \centering
    \includegraphics[width=\linewidth]{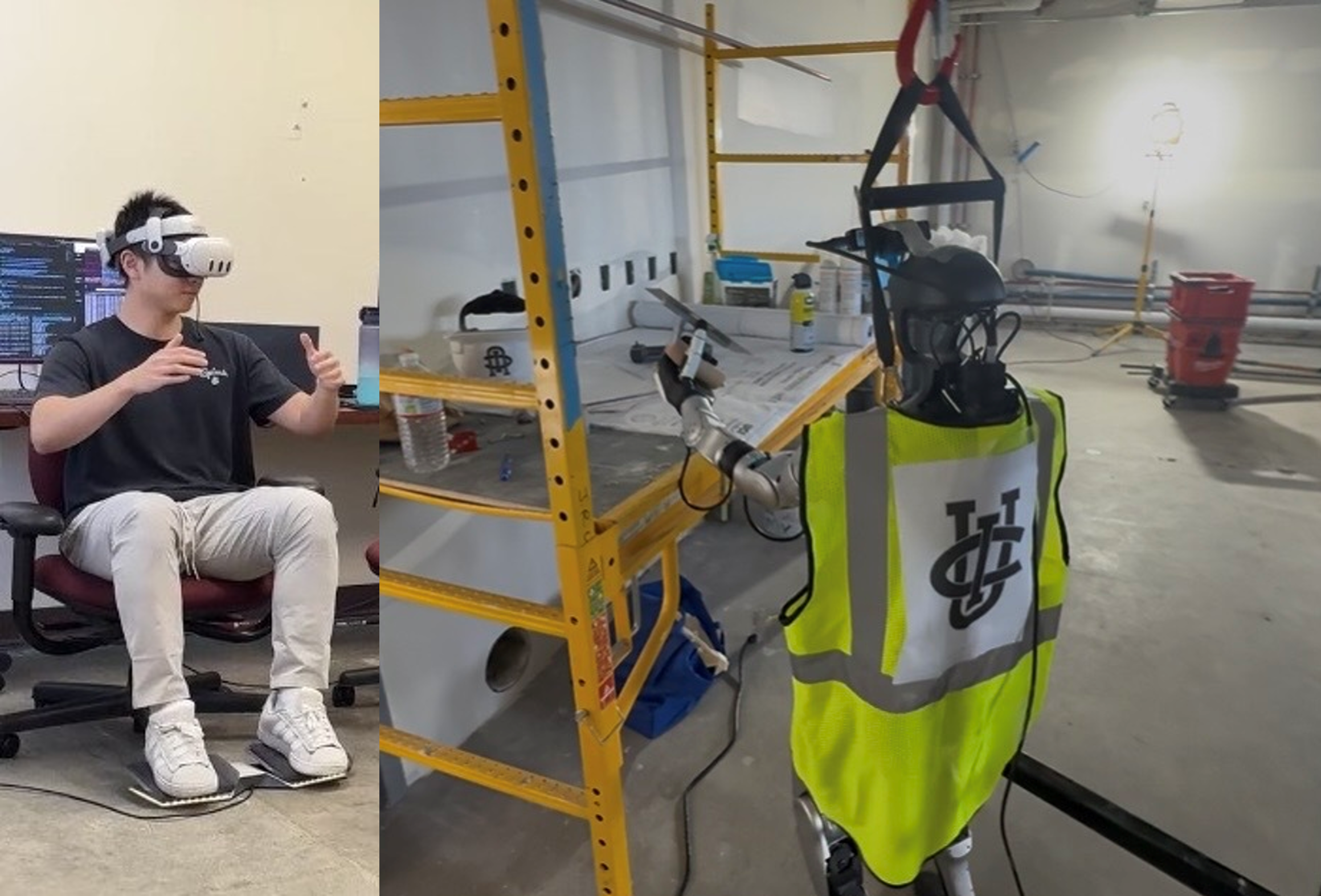}
    \caption{Humanoid teleoperation for construction task. (Left) An operator controls the robot using a Meta Quest 3 headset and a GYLDR foot controller. (Right) A teleoperated Unitree G1 performs a tool transport task on a job site. It grasps a hand tool (a trowel) from the work surface and carries it toward the destination container (red box).}
    \label{fig1}
\end{figure}

Humanoids are a natural fit for on-site construction work because their morphology matches spaces and tools built for humans. They can navigate tight spaces, climb steps, and use common tools without site modification or task-specific automation \cite{c1}. This allows humanoids to address a broad range of tasks with a single embodiment. While the existing humanoid models may not be optimal for every task, they offer reach and mobility no fixed or wheeled system can match \cite{c1}. However, reliable, full autonomy of humanoid robots remains out of the reach in the noise and clutter of a real construction site \cite{c1}.

Teleoperation offers a practical near-term path: skilled operators control robots remotely, reducing physical strain and hazard to workers while generating high-quality demonstration data for training future autonomous models. Manual material handling is a leading cause of musculoskeletal disorders \cite{c4}, and over 1,000 construction workers die on the job each year in the United States \cite{c5}. These conditions compound a persistent labor shortage, as an aging workforce approaches retirement while the physical toll deters new entrants \cite{c6}. By transferring strain and hazard to the robot while leveraging the operator's expertise, teleoperation can improve worker safety and career longevity. Moreover, recent whole-body teleoperation systems have shown that expert demonstrations are key to training autonomous vision-language-action (VLA) models \cite{c7,c8,c9}. However, existing whole-body teleoperation systems have been evaluated primarily on household and laboratory tasks, leaving their effectiveness for construction work largely unexplored. 

In this work, we develop a teleoperation system for a humanoid construction robot from a seated workstation using a Meta Quest 3 extended-reality (XR) headset for head and hand tracking and GLYDR foot pedals for locomotion control. We then deploy a Unitree G1 humanoid robot on an active construction site and perform a feasibility study on two high-importance construction tasks selected from the O*NET occupational database \cite{onet}. Our key contributions are:

\begin{figure*}[t]
    \centering
    \includegraphics[width=0.8\linewidth]{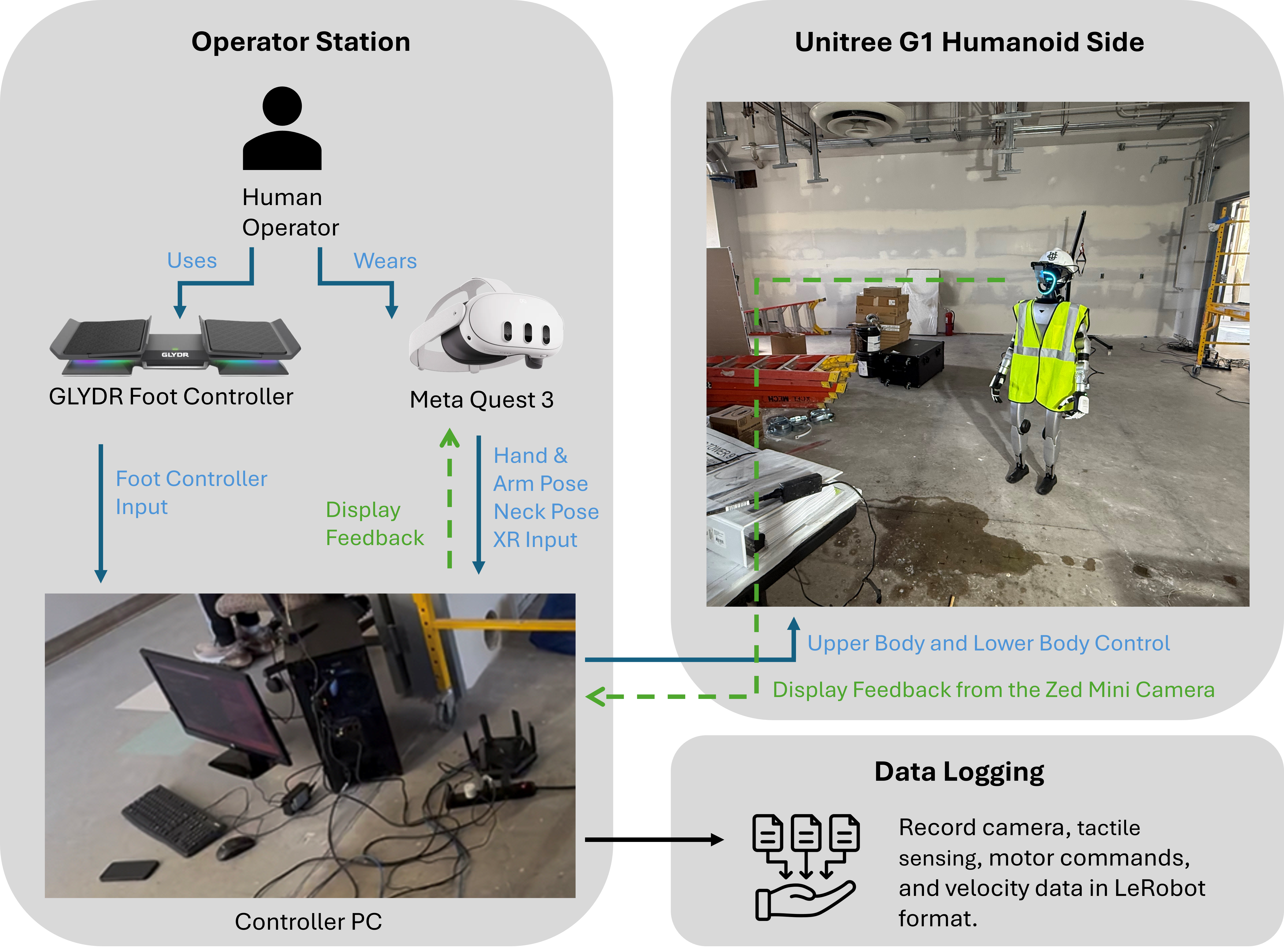}
    \caption{Diagram of the teleoperation system architecture. The operator controls the Unitree G1 humanoid using a Meta Quest 3 headset for upper-body and neck motion and GLYDR foot controller for locomotion. The control inputs are processed through a controller PC, while egocentric video from the robot-mounted ZED Mini camera is streamed to the operator. All camera, tactile, motor commands, and velocity data are recorded in LeRobot format to support future policy learning. }
    \label{fig:method}
\end{figure*}

\begin{itemize}
    \item A humanoid robot teleoperation system allowing seated, remote operation for simultaneous manipulation and locomotion built using open source components;
    \item Deployment on an active construction site;
    \item A preliminary evaluation of teleoperation performance on two representative construction tasks.
\end{itemize}

\section{RELATED WORK}
Humanoid teleoperation systems span a spectrum of control coupling. Decoupled systems control the upper and lower body through separate interfaces \cite{c11,c12}, while recent whole-body systems track the operator's full pose to drive the entire body \cite{c7,c8,c9}. These systems have enabled humanoids to perform diverse real-world tasks, including locomotion \cite{c7,c8,c9}, object manipulation and transport \cite{c7,c8,c9}, and even dexterous medical procedures \cite{c14,invivo}.  Many existing systems \cite{c8, c9} also require the operator to physically reproduce the robot's motions, including walking comparable distances and squatting to reach low objects retaining much of the physical burden and limiting the feasibility of sustained operation. 

Construction robotics has pursued task-specific automation \cite{brosque} \cite{onsite} as well as autonomous humanoids such as HRP-5P, which demonstrated drywall installation but relies on extensive environmental modeling and remains slow outside scripted conditions \cite{c15}. More recently, teleoperation has been applied to construction arms for precise pick-and-place tasks while collecting imitation-learning data \cite{c16}, as well as semi-humanoid mobile manipulators for construction tasks \cite{c17}. However, to the best of our knowledge, no prior work has evaluated the feasibility of full body teleoperation of a humanoid robot for construction tasks.

\section{METHODS}

\subsection{System Architecture}
The teleoperation system, illustrated in Figure~\ref{fig:method}, is designed to enable remote, seated operation of a humanoid robot on an active job site. At the operator workstation, the operator wears a Meta Quest 3 headset for egocentric video streaming and hand tracking, and uses a GLYDR foot controller for lower-body locomotion. These devices connect to the controller PC, which relays commands from the GLYDR and Quest 3 to a Unitree G1 humanoid equipped with Inspire five-finger hands with tactile sensing. Together, these components achieve three objectives: (1) an extended-reality perception system that lets the operator actively look around from the robot's point of view, (2) hand tracking based manipulation, and (3) hands-free locomotion control that enables simultaneous locomotion and manipulation control by a seated operator.

\subsubsection{Active Visual Perception}
To enable egocentric, active perception, the robot is equipped with a StereoLabs ZED Mini stereo camera mounted on the 2-degrees of freedom (DoF) neck developed in \cite{c9}. The stereo images are streamed to the Quest headset via WebRTC. The operator's head pitch and yaw, recorded from the Quest headset, are converted to pulse width modulation (PWM) values that drive the camera to match the operator's head orientation.

\subsubsection{Hand Tracking} The robot's upper body and hands are controlled through Unitree’s XR teleoperate framework \cite{xr_teleoperate} which maps operator and and finger poses to robot joint positions. While the hand tracking provides intuitive, precise control of the robot hands, it necessitates that hands free controllers are used for the rest of the teleoperation system.

\subsubsection{Hands Free Locomotion} To enable simultaneous locomotion and manipulation (e.g., walking while carrying or operating a tool), the robot's locomotion is controlled using GLYDR foot pedals. The GLYDR interface consists of independent left and right foot pedals that provide differential control. With both pedals in the neutral position, the robot remains stationary. Moving either pedal forward commands a turn in the opposite direction, while moving both pedals forward simultaneously commands forward locomotion. The resulting linear and angular velocity commands are executed by GR00T Whole Body Control \cite{GR00T}, which generates the corresponding lower-body motor commands. Auxiliary flaps on the GLYDR pedals control squatting and standing through GR00T Whole Body Control, enabling hands-free height adjustment for manipulation near the ground or on lower shelves.

\subsection{Data Collection}
We implemented a data collection pipeline that records each teleoperation session in the LeRobot dataset format, capturing the Zed Mini Stereo camera stream, tactile sensing, motor states, motor actions, and velocity commands. The resulting synchronized demonstrations can be used as training data for imitation-learning, allowing each teleoperated task to both accomplish useful work and contribute data toward future autonomous execution.

\section{EXPERIMENTS}
We evaluated the system on two representative high-importance construction tasks (Table~\ref{tab:ONET}) drawn from the O*NET occupation database \cite{onet} for Construction Laborers, that fall within the platform's payload and grasp limits.

\textbf{Task 1: Tool Transport}. Representing the first task type in Table~\ref{tab:ONET}, the operator controls the robot to grasp a misplaced hand tool, walk approximately 4 meters to a designated tool bin, and place the tool inside (Figure~\ref{fig1}). This task evaluates simultaneous locomotion and manipulation through the GLYDR pedals and Quest hand tracking. The success rate for this task was defined by the successful pickup and transportation of the desired tool to the target location.

\textbf{Task 2: Painting}. Representing the second task type in in Table~\ref{tab:ONET}, the operator uses the robot to grasp a roller brush and apply a paint coating across a piece of paper. This task evaluates sustained, continuous-contact manipulation, requiring the operator to maintain contact and consistent coverage during an extended arm motion. The success rate for this task was defined by whether the paint coating filled at least 90\% of the paper.

\begin{table}
    \centering
    \caption{Selected Tasks from O*NET Occupation Specific Tasks for Construction Laborers \cite{onet}}
    \label{tab:ONET}
    \begin{tabular}{cc p{0.60\columnwidth}}
    \hline
       Task  & O*NET Category & Description \\
       \hline
       1  & Core & Load, unload, or identify building materials, machinery, or tools, distributing them to the appropriate locations, according to project plans or specifications. \\
       2  & Supplemental & Mop, brush, or spread paints, cleaning solutions, or other compounds over surfaces to clean them or to provide protection. \\
         \hline
    \end{tabular}
\end{table}

\begin{table*}[t]
    \centering
    \caption{Teleoperation task performance across two representative tasks.}
    \label{tab:task_results}
    \begin{tabular}{lccccc}
        \hline
        \textbf{Task} & \textbf{Operator} & \textbf{Teleoperation Success Rate (\%)} & \textbf{Avg. Teleoperation Time (s)} & \textbf{Avg. Baseline Time (s)} &\textbf{Trials} \\
        \hline
        Tool Transport & VR-novice & 100 & 83 & 5 & 10 \\
        Tool Transport & VR-expert & 100 & 62 & 3 & 10 \\        
        \multicolumn{1}{l}{\textit{\textbf{Tool Transport}}} & 
        \textit{\textbf{Avg}} & \textit{\textbf{100}} & \textit{\textbf{72}} & \textit{\textbf{4}} & \textit{\textbf{20}} \\      
        Painting & VR-novice & 80 & 169 & 40 & 10 \\
        Painting & VR-expert & 80 & 129 & 53 & 10 \\        
        \multicolumn{1}{l}{\textit{\textbf{Painting}}} & 
        \textit{\textbf{Avg}} & \textit{\textbf{80}} & \textit{\textbf{149}} & \textit{\textbf{47}} & \textit{\textbf{20}} \\      
        \hline
    \end{tabular}
\end{table*}

Table~\ref{tab:task_results} reports both the success rate and the average time required to complete each task. Each task was performed by two operators, one with prior experience using virtual reality (VR) interfaces and one without, over 10 trials each. Additionally, the time it took each operator to perform the task manually without the robot was recorded as a baseline.  Both operators were able to complete tool transport with a 100\% success rate and painting with an 80\% success rate, demonstrating the feasibility of using a teleoperated humanoid for construction tasks. However, teleoperated tasks took substantially longer than manual execution. Tool Transport averaged approximately 72 seconds versus 4 seconds manually ($\sim$18$\times$ slowdown), while painting took 149 seconds via teleoperation compared with 47 seconds manually ($\sim$3$\times$ slowdown). The larger penalty for tool transport is consistent with the additional locomotion and repositioning required to carry an object between locations, whereas painting primarily involved stationary, continuous contact manipulation with minor positional adjustments, such as moving forward or turning as needed. VR experience also appeared to influence teleoperation efficiency. with the operator experienced with VR headsets completing both tasks faster than the novice. 

The lower success rate for painting highlights the challenges of sustained-contact manipulation. Painting failures were primarily associated with unsuccessful or unstable grasp of the roller brush. This suggests that reliable grasp acquisition and retention are important limitations of pure hand tracking systems. Beyond grasp stability, we also observed motor overheating during extended teleoperation sessions, highlighting the need for improved control strategies to support sustained operation and practical deployment.

\section{CONCLUSION}
This work presented a teleoperation system for construction tasks using a Unitree G1 humanoid, combining XR-based upper-body control with pedal-based locomotion, and evaluated it on two construction tasks, demonstrating the potential of teleoperated humanoids for construction applications. Preliminary results demonstrate high success rates on both tasks (100\% and 80\% respectively), indicating the ability of the system to simultaneously accomplish meaningful work while generating valuable training data. However, feasibility testing revealed important limitations. Teleoperation was much slower than manual task completion as grasp stability posed a key challenge to the painting task. We also observed motor overheating during extended teleoperation, highlighting continuous operation as an important practical challenge for deployment of humanoid systems.

Future work will focus on expanding the task suite to a wide variety of construction tasks covering tool use, site clean up, and assembly tasks on diverse jobsites. The collected demonstration dataset will be published to assist in training autonomous policies. Additionally, future work will focus on improving the effectiveness of the teleoperation system through haptic feedback methods for the operator and control approaches that reduce motor overheating. These efforts aim to bridge the gap between the demonstrated feasiblity of humanoid teleoperation indicated by this study and reliable, efficient construction work.


\addtolength{\textheight}{-12cm}   




\section*{ACKNOWLEDGMENT}
We would like to thank Roy Afaryan, Shaurya Sharma, and Javier Ramos for their contributions to the development and testing of the teleoperation system.


\end{document}